\documentclass{article}

\def\E{\mathop{\mathbb{E}}} % redef to make it an operator

\def\cN{\mathcal{N}}

\def\cR{\mathcal{R}}

\newcommand\indep{\protect\mathpalette{\protect\independenT}{\perp}}
\def\independenT#1#2{\mathrel{\rlap{$#1#2$}\mkern2mu{#1#2}}}

\usepackage{charter}
\usepackage{amsmath}
\usepackage[usenames,dvipsnames]{xcolor}
\usepackage{graphicx}
\usepackage[labelfont=bf]{caption}
\usepackage{booktabs}

\usepackage{listings}
\usepackage{fancyvrb}
\fvset{fontsize=\normalsize}

\usepackage[acronym,shortcuts,smallcaps,nowarn,nohypertypes={acronym,notation}]{glossaries}

\lstdefinestyle{mystyle}{
    commentstyle=\color{OliveGreen},
    keywordstyle=\color{BurntOrange},
    numberstyle=\tiny\color{black!60},
    stringstyle=\color{MidnightBlue},
    basicstyle=\ttfamily,
    breakatwhitespace=false,
    breaklines=true,
    captionpos=b,
    keepspaces=true,
    numbers=left,
    numbersep=5pt,
    showspaces=false,
    showstringspaces=false,
    showtabs=false,
    tabsize=2
}
\usepackage[colorlinks,linktoc=all]{hyperref}
\usepackage[all]{hypcap}
\hypersetup{citecolor=MidnightBlue}
\hypersetup{linkcolor=MidnightBlue}
\hypersetup{urlcolor=MidnightBlue}

\usepackage[nameinlink]{cleveref}

\usepackage{tikz}
\usetikzlibrary{shapes,decorations,arrows,calc,arrows.meta,fit,positioning}
\tikzset{
    -Latex,auto,node distance =1 cm and 1 cm,semithick,
    state/.style ={circle, draw, minimum width = 0.7 cm},
    detstate/.style ={rectangle, draw, minimum width = 0.7 cm, minimum height = 0.7 cm},
    point/.style = {circle, draw, inner sep=0.04cm,fill,node contents={}},
    bidirected/.style={Latex-Latex,dashed},
    el/.style = {inner sep=2pt, align=left, sloped}
}

\usepackage{wrapfig}
\usepackage{mathtools}

\newacronym{KL}{kl}{Kullback-Leibler}
\newacronym{ELBO}{elbo}{\emph{evidence lower bound}}
\newacronym{POPELBO}{pop-elbo}{\emph{population evidence lower bound}}

\newacronym{SVI}{svi}{stochastic variational inference}
\newacronym{nll}{nll}{negative log likelihood}
\newacronym{conc}{conc}{concordance}

\newacronym{svm}{svm}{svm}

\newacronym{BUMPVI}{bump-vi}{bumping variational inference}

\newacronym{GMM}{gmm}{Gaussian mixture model}
\newacronym{LDA}{lda}{latent Dirichlet allocation}

\newacronym{SUTVA}{sutva}{stable unit treatment value assumption}

\newacronym{KSD}{ksd}{{kernelized Stein discrepancy}}
\newacronym{KCC-SD}{kcc-sd}{kernelized complete conditional Stein discrepancy}

\newacronym{OPVI}{opvi}{operator variational inference}
\newacronym{SVGD}{svgd}{Stein variational gradient descent}

\newacronym{vde}{vde}{variational decoupling}
\newacronym{cfn}{cfn}{control-function method}
\newacronym{gcfn}{gcfn}{generalized control-function method}
\newacronym{2sls}{2sls}{two-stage least-squares method}
\newacronym{gmm}{gmm}{generalized method of moments}
\newacronym{iv}{iv}{instrumental variable}
\newacronym{cdf}{cdf}{cumulative distribution function}

\newacronym{ours}{x-cal}{explicit calibration}
\newcommand{\xcaled}{\textsc{x-calibrated}}
\newacronym{d-cal}{d-calibration}{distributional calibration}
\newacronym{d-cal-short}{d-cal}{d-cal}
\newacronym{crps}{crps}{continuous ranked probability score}
\newacronym{s-crps}{s-crps}{Survival-\acrshort{crps}}
\newacronym{ifd}{ifd}{individual failure distribution}

\newacronym{hl}{hl}{Hosmer-Lemeshow}
\newacronym{gb}{gb}{Grønnesby-Borgan}
\newacronym{dn}{dn}{D’Agostino-Nam}
\newacronym{km}{km}{Kaplan-Meier}

\newacronym{ni}{ni}{Not-Interpolated}
\newacronym{i}{i}{Interpolated}
\newacronym{mimic-iii}{mimic-iii}{Medical Information Mart for Intensive Care}
\newacronym{mnist}{mnist}{Modified National Institute of Standards and Technology database}
\newacronym{tcga}{tcga}{The Cancer Genome Atlas}
\newacronym{mtlr}{mtlr}{Multi-Task Logistic Regression}
\newacronym{aft}{aft}{Accelerated Failure Times}

\DeclareRobustCommand{\mb}[1]{\ensuremath{\boldsymbol{\mathbf{#1}}}}

\renewcommand{\mid}{~\vert~}

\newcommand{\mbc}{\mb{c}}
\newcommand{\mbd}{\mb{d}}

\newcommand{\mbt}{\mb{t}}
\newcommand{\mbu}{\mb{u}}

\newcommand{\mbx}{\mb{x}}
\newcommand{\mby}{\mb{y}}
\newcommand{\mbz}{\mb{z}}

\newcommand{\vx}{x}

\newcommand{\vt}{t}

\newcommand{\mbmu}{\mb{\mu}}

\newcommand{\cI}{\mathcal{I}}

\newcommand{\g}{\mid}

\crefname{lemma}{lemma}{lemmas}
\crefname{prop}{proposition}{propositions}

\DeclareRobustCommand{\indicator}[1]{\ensuremath{\mathbbm{1}\left[#1\right]}}
\usepackage{amsfonts} 
\usepackage{amsthm}

\theoremstyle{plain}

\theoremstyle{definition}
\newtheorem*{assumption*}{Assumption}

\usepackage{booktabs}
\usepackage{bbm}
\usepackage{nicefrac}
\usepackage{enumitem}
\usepackage{siunitx}
\usepackage{soul}

    \usepackage[final]{neurips_2020}

\usepackage[utf8]{inputenc}
\usepackage[T1]{fontenc}
\usepackage{url}
\usepackage{booktabs}
\usepackage{amsfonts}
\usepackage{nicefrac}
\usepackage{microtype}

\title{X-CAL: Explicit Calibration for Survival Analysis}

\author{
 Mark Goldstein\thanks{Equal Contribution} \\
  New York University\\
  \texttt{goldstein@nyu.edu} \\
  \And
 Xintian Han\footnotemark[1]\\
  New York University\\
  \texttt{xintian.han@nyu.edu} \\
    \And 
 Aahlad Puli\footnotemark[1]\\
  New York University\\
  \texttt{aahlad@nyu.edu} \\
   \And 
    Adler J. Perotte\\
 Columbia University\\
  \texttt{adler.perotte@columbia.edu} \\
   \And 
    Rajesh Ranganath\\
  New York University\\
  \texttt{rajeshr@cims.nyu.edu} \\
  }
  
\begin{document}

\maketitle

\begin{abstract}

Survival analysis models the distribution of time until an event of interest, such as discharge from the hospital or admission to the ICU. When a model's predicted number of events within any time interval is similar to the observed number, it is called \textit{well-calibrated}. A survival model's calibration can be measured using, for instance, \gls{d-cal} \citep{haider2020effective} which computes the squared difference between the observed and predicted number of events within different time intervals.  Classically, calibration is addressed in post-training analysis. 
We develop \gls{ours}, which turns \gls{d-cal} into a differentiable objective that can be used in survival modeling alongside maximum likelihood estimation and other objectives. \Gls{ours} allows practitioners to directly optimize calibration and strike a desired balance between predictive power and calibration.
In our experiments, we fit a variety of shallow and deep models
on simulated data, a survival dataset based on \acrshort{mnist}, on length-of-stay prediction using \acrshort{mimic-iii} data, and on brain cancer data from The Cancer Genome Atlas. We show that the models we study can be miscalibrated. We give experimental evidence on these datasets that \gls{ours} improves \gls{d-cal} without a large decrease in concordance or likelihood. 

\end{abstract}

\section{Introduction}

A core challenge in healthcare
 is to assess the risk of events such
 as onset of disease or death. Given a patient's vitals and lab values, physicians should know whether the patient is at risk for transfer to a higher level of care.
 Accurate estimates of the time-until-event help physicians assess risk and accordingly prescribe treatment strategies:
  doctors match aggressiveness of treatment against severity of illness. These predictions are important to the health of the individual patient and to the allocation of resources in the healthcare system, affecting all patients.
 
Survival Analysis
formalizes this risk
assessment by estimating the conditional distribution of the \textit{time-until-event} for an outcome of interest, called the failure time. Unlike supervised learning, survival analysis must handle datapoints that are \textit{censored}: their failure time is not observed, but bounds on the failure time are. For example, in a 10 year cardiac health study  \citep{wilson1998prediction,d2008general}, some individuals will remain healthy over the study duration. 
Censored points are informative, as we can learn that
 someone's physiology indicates they are healthy-enough to avoid
onset of cardiac issues within the next 10 years.

A \textit{well-calibrated} survival model is one where the predicted number of events within any time interval is similar to the observed number \citep{pepe2013methods}. When this is the case, 
event probabilities can be interpreted as risk and can be used for downstream tasks, treatment strategy, and  
human-computable risk score development \citep{sullivan2004presentation,demler2015tests,haider2020effective}.
Calibrated conditional models enable
accurate, individualized prognosis and may help prevent giving patients misinformed limits on their survival, such as 6 months when they would survive years. Poorly calibrated predictions of time-to-event can misinform decisions about a patient's future.

Calibration is a concern in today's deep models.
Classical neural networks that were not wide or deep by modern standards
were found to be as calibrated
as other models after the latter were calibrated (boosted trees, random forests, and \acrshortpl{svm} calibrated using Platt scaling and isotonic regression) \citep{niculescu2005predicting}.
However, deeper and wider models using batchnorm and dropout have been found to be overconfident or otherwise miscalibrated
\citep{guo2017calibration}.
Common shallow survival models such as the Weibull \gls{aft} model may also be miscalibrated \citep{haider2020effective}. We explore shallow and deep models in this work.

Calibration checks are usually performed post-training. This approach decouples the search for a good predictive model and a well-calibrated one~\citep{song2019distribution,platt1999probabilistic,zadrozny2002transforming}.
Recent approaches tackle calibration in-training via alternate loss functions. However, these may not,
even implicitly, optimize a well-defined calibration measure, nor do they allow for explicit balance between prediction and calibration
 \citep{avati87countdown}.
Calibration during training has been explored
recently  for binary classification 
\citep{kumar2018trainable}.
Limited evaluations of calibration in survival models can be done by considering only particular time points: \textit{this model is well-calibrated for half-year predictions}. Recent work considers \gls{d-cal} \citep{haider2020effective}, a holistic measure of calibration of time-until-event that measures calibration of \textit{distributions}.

In this work, we propose to improve calibration by augmenting traditional objectives for survival modeling with a differentiable approximation of \gls{d-cal},  which we call \glsreset{ours}\gls{ours}.
\Gls{ours} is a plug-in objective that reduces obtaining good calibration to an optimization problem amenable to data sub-sampling.
\Gls{ours} helps build well-calibrated versions of many existing models and controls calibration \textit{during} training.
In our experiments \footnote{Code is available at \href{https://github.com/rajesh-lab/X-CAL}{https://github.com/rajesh-lab/X-CAL}}, we fit a variety of shallow and deep models
on simulated data, a survival dataset based on \acrshort{mnist}, on length-of-stay prediction using \acrshort{mimic-iii} data, and on brain cancer data from
The Cancer Genome Atlas. We show that the models we study can be miscalibrated. We give experimental evidence on these datasets that \gls{ours} improves \gls{d-cal} without a large decrease in concordance or likelihood. 

\section{Defining and Evaluating Calibration in Survival Analysis}

Survival analysis models the time $\mbt > 0$
until an event, called the failure time.
$\mbt$ is often assumed to be conditionally distributed
given covariates $\mbx$. Unlike typical regression problems, there may also be censoring times $\mbc$ that determine
whether $\mbt$ is observed.
We focus on right-censoring in this work, 
with observations $(u,\delta,x)$ where $\mbu=\min(\mbt,\mbc)$ and $\boldsymbol{\delta}=\indicator{\mbt < \mbc}$.
If $\delta=1$ then $u$ is a failure time. Otherwise $u$ is a censoring time and the datapoint is called \textit{censored}. Censoring times may be constant or random. We assume censoring-at-random: $\mbt \indep \mbc \mid \mbx$.

We denote the joint distribution of $(\mbt,\mbx)$ by $P$ and the conditional \gls{cdf}
of $\mbt \g \mbx$ by $F$ (sometimes denoting the marginal \gls{cdf} by $F$ when clear). Whenever distributions or \glspl{cdf} have no subscript parameters, they are taken to be true data-generating distributions and when they have parameters $\theta$ they denote a model.
We give more review of key concepts, definitions, and common survival analysis models in \Cref{appsec:appendixbackground}.

\subsection{Defining Calibration} 
We first establish a common definition of calibration for binary outcome. Let $\mbx$ be covariates and
let $\mbd$ be a binary outcome distributed conditional on $\mbx$.
Let them have joint distribution $P(\mbd,\mbx)$. Define $\mathtt{risk}_\theta(\vx)$ as the modeled probability $P_\theta(\mbd=1 \g \vx)$, a deterministic function of $\vx$.
\cite{pepe2013methods} define calibration as the condition that
\begin{align}
\label{eq:binarycalibration}
    \mathbb{P}(\mb{d} = 1 \g  \mathtt{risk}_\theta(\mbx)=r)
    = 
    \approx r.
\end{align}
That is, the frequency of events is $r$ among subjects whose modeled risks are equal to $r$. For a survival problem with joint distribution
$P(\mbt,\mbx)$, we can define risk
to depend on an observed failure time instead of the binary outcome $\mbd=1$. With $F_\theta$ as the model \gls{cdf}, the
definition of risk for survival analysis becomes
 $\mathtt{risk}_{\theta}(\vt, \vx) = F_\theta(\vt\g \vx) $, a deterministic function of $(\vt,\vx)$.
% For $\mbt, \mbx \sim $
Then perfect calibration is the condition that,
for all sub-intervals $I=[a,b]$ of $[0,1]$,
\begin{align}
\label{eq:weakcalibration}
   \mathbb{P}(\mathtt{risk}_\theta(\mbt, \mbx) \in  I) 
       = \E_{P(\mbt, \mbx)} \indicator{ F_\theta(\mbt \g \mbx) \in  I} = |I|.
\end{align}
This is because, for continuous $F$ (an assumption we keep for the remainder of the text), \glspl{cdf} transform samples of their own distribution to $\text{Unif}(0,1)$ variates. Thus, when model predictions are perfect and $F_\theta=F$, the probability that $F_\theta(\mbt \g \mbx)$ takes a value in interval $I$ is equal to $|I|$. Since the expectation is taken over $\mbx$, the
same holds when $F_\theta(\vt \g \vx) = F(\vt)$, the true marginal \gls{cdf}.

\subsection{Evaluating Calibration}
\label{sec:evalcalibration}

Classical tests and their recent modifications assess calibration of survival models for a particular time of interest $\vt^*$ by comparing observed versus modeled event frequencies \citep{lemeshow1982review,gronnesby1996method,d2003evaluation,royston2013external,demler2015tests,yadlowsky2019calibration}. They apply the condition in
\Cref{eq:binarycalibration} for the classification task $\mbt < \vt^* \g \mbx$. These tests are limited in two ways 1) it is not clear how to combine calibration assessments over the entire range of possible time predictions \citep{haider2020effective} and 2) they answer calibration in a  rigid yes/no fashion with hypothesis testing. We briefly review these tests in
\Cref{appsec:appendixbackground}. 

\paragraph{\Gls{d-cal}} \cite{haider2020effective} develop \glsreset{d-cal}\gls{d-cal} to test the calibration of conditional survival \textit{distributions} across all times. \Gls{d-cal} uses the
condition in \Cref{eq:weakcalibration}
and checks the extent to which it
holds by evaluating the model conditional
\gls{cdf} on times in the data
and checking 
that these \gls{cdf} evaluations are uniform over $[0,1]$. This uniformity ensures that observed and predicted numbers of events within each time interval match.

To set this up formally, recall that $F$ denotes the unknown true \gls{cdf}.
For each individual $\vx$, let $F_\theta(\mbt \g \vx)$ denote the modeled \gls{cdf} of time-until-failure.
To measure overall calibration error,  \gls{d-cal} accumulates the squared errors of the equality condition in \Cref{eq:weakcalibration} over sets $I\in \cI$ that cover $[0,1]$:
\begin{align}\label{eq:dcal}
    \cR(\theta):= \sum_{I\in \cI} \left(\E_{P(\mbt, \mbx)} \indicator{F_{\theta}(\mbt \g \mbx) \in I} - |I|\right)^2.
\end{align}
The collection $\cI$ is chosen to contain disjoint contiguous intervals $I \subseteq [0,1]$, that cover the whole interval $[0,1]$. \cite{haider2020effective} perform a $\chi^2$-test to determine whether a model is well-calibrated,
replacing the expectation in \Cref{eq:dcal} with a Monte Carlo estimate. 

\paragraph{Properties}
Setting aside the hypothesis testing step, we highlight two key properties of \gls{d-cal}. First, \gls{d-cal} is zero for the correct conditional model. 
This ensures that the correct model is not wrongly mischaracterized as miscalibrated.
Second, for a given model class and dataset, smaller \gls{d-cal} means a model is more calibrated. This means that it makes sense to minimize \gls{d-cal}.
Next, we make use of these properties and turn \gls{d-cal} into a differentiable objective.

\section{ \Gls{ours}: A Differentiable Calibration Objective \label{sec:xcal}}
We measure calibration error with
\gls{d-cal} (\Cref{eq:dcal}) and propose to incorporate it into our training and minimize it directly.
However, the indicator function $\indicator{\cdot}$ poses a challenge for optimization.
Instead, we derive a soft version of \gls{d-cal} using a soft set membership function. We then develop an upper-bound to
soft \gls{d-cal} that we call
\gls{ours}
that supports subsampling for stochastic optimization with batch data.

\subsection{Soft Membership \gls{d-cal}}
We replace the membership indicator for a set $I$ with a differentiable function.
Let $\gamma > 0$ be a temperature parameter. Let $\sigma(x)=(1+ \exp[-x])^{-1}$.
For point $u$ and the set $I = [a, b]$, define soft membership $\zeta_\gamma$ as
\begin{align}
\label{eq:zeta}
\zeta_\gamma(u; I) :=  \sigma(\gamma(u-a)(b-u)),
\end{align}
where $\gamma\rightarrow\infty$ makes membership exact.
This is visualized in \Cref{fig:choiceofgamma} 
in \Cref{appsec:gammahyperparameter}.
We propose the following differentiable approximation to
\Cref{eq:dcal}, which we call
soft \gls{d-cal},
for use in a calibration objective:
\begin{align}\label{eq:soft-d-cal}
    \hat{\cR}_\gamma(\theta) := \sum_{I\in \cI} \left(\E_{P(\mbt, \mbx)} \zeta_\gamma \left(F_{\theta}(\mbt\g \mbx);I \right) - |I|\right)^2.
\end{align}
We find that $\gamma=10^4$ allows
for close-enough approximation to optimize exact \gls{d-cal}.

\subsection{Stochastic Optimization via Jensen's Inequality}

Soft \gls{d-cal} squares an expectation over the data, meaning that its gradient includes a product of two expectations over the same data. Due to this,
it is hard to obtain a low-variance, unbiased gradient estimate with batches of data, which is important for models that rely on stochastic optimization.
To remedy this, we develop an upper-bound on soft \gls{d-cal}, which we call \gls{ours}, whose gradient has an easier unbiased estimator. 

Let $R_{\gamma,\theta}(\vt,\vx,I)$ denote the contribution to soft \gls{d-cal} error due to one set $I$ and a single sample $(\vt,\vx)$ in \Cref{eq:soft-d-cal}:
$R_{\gamma,\theta}(\vt, \vx, I) :=  \zeta_\gamma \left(F_{\theta}(\vt\g \vx);I \right) - |I|$.
Then soft \gls{d-cal} can be written as:
\begin{align*}
\hat{\cR}_\gamma(\theta) = \sum_{I\in \cI} \left(\E_{P(\mbt, \mbx)} R_{\gamma,\theta}(\mbt, \mbx, I)\right)^2.
\end{align*}

For each term in the sum over sets $I$, we proceed by in two steps. First, replace the expectation over data
$\E_{P} $ with an expectation over sets of samples $\E_{S\sim P^M}$ of the mean of $R_{\gamma,\theta}$ where $S$ is a set of size $M$. Second, use Jensen's inequality
to switch the expectation and square.
\begin{align}\label{eq:upper-bound}
    \begin{split}
    \hat{\cR}_\gamma(\theta) = \sum_{I\in \cI}
 \left(\E_{S\sim P^M}\frac{1}{M}\sum_{\vt,\vx \in S} R_{\gamma,\theta}(\vt, \vx, I)\right)^2 \leq  \E_{S\sim P^M} 
 \sum_{I\in \cI} \left(\frac{1}{M}\sum_{\vt,\vx \in S}R_{\gamma,\theta}(\vt, \vx, I)\right)^2.
    \end{split}
\end{align}
We call this upper-bound \gls{ours}
and denote it by $\hat{\cR}_\gamma^+(\theta)$.
To summarize, $\lim_{\gamma \rightarrow \infty} \hat{\cR}_\gamma(\theta) = \cR(\theta)$
by soft indicator approximation and $\hat{\cR}_\gamma(\theta) \leq \hat{\cR}_\gamma^+(\theta)$ by Jensen's inequality.
As $M \rightarrow \infty$, the slack introduced due to Jensen's inequality vanishes (in practice we are constrained by the size of the dataset).
We now derive the gradient with respect to $\theta$, using $\zeta^\prime(u) = \frac{d \zeta}{ d u }(u)$:
\begin{align}\label{eq:gradient-of-upper-bound}
    \frac{d \hat{\cR}_\gamma^+}{d \theta} = \E_{S\sim P^M} \sum_{I\in \cI} \frac{2}{M^2}\sum_{\vt,\vx \in S} R_{\gamma,\theta}(\vt, \vx, I)
    \left( \zeta_\gamma^\prime \left({ F_{\theta}}(\vt\g \vx);I\right)  \frac{ d F_{\theta}}{ d\theta}(\vt\g \vx) \right).
\end{align}
We estimate \Cref{eq:gradient-of-upper-bound} by sampling batches $S$ of size $M$ from the empirical data. 

Analyzing this gradient demonstrates how \gls{ours} works. If the fraction of points in bin $I$ is larger than $|I|$, \gls{ours} pushes points out of $I$. The gradient of $\zeta_\gamma$ pushes points in the first half of the bin to have smaller \gls{cdf} values and similarly points in the second half are pushed upwards. 

While this works well for intervals not at the boundary of $[0,1]$, some care must be taken at the boundaries. \gls{cdf} values in the last bin may be pushed to one and unable to leave the bin. Since the maximum \gls{cdf} value is one, $\indicator{u \in [a,1]} = \indicator{u \in [a,b]}$ for any $b>1$. Making use of this property, \gls{ours} extends the right endpoint of the last bin so that all \gls{cdf} values are in the first half of the bin and therefore are pushed to be smaller. The boundary condition near zero is similar.
We provide further analysis in \Cref{appsec:mod-soft-ind}.

\gls{ours} can be added to loss functions such as \gls{nll} and other survival modeling objectives such as \gls{s-crps} \citep{avati87countdown}. For example, the full \xcaled{} maximum likelihood objective for a model $P_\theta$ and $\lambda > 0$ is:
\begin{align}
    \min_\theta  \E_{P(\mbt, \mbx)} -\log P_\theta(\mbt \g \mbx) + \lambda \hat{\cR}_\gamma^+(\theta).
\end{align}

\paragraph{Choosing $\gamma$}
For small $\gamma$, soft \gls{d-cal}
is a poor approximation to \gls{d-cal}. For large $\gamma$, gradients vanish,
making it hard to optimize \gls{d-cal}.
We find that setting $\gamma=10000$ worked in all experiments.
We evaluate the choice of $\gamma$ in 
\Cref{appsec:gammahyperparameter}. 

\paragraph{Bound Tightness} The slack in Jensen's inequality does not adversely affect our experiments in practice. We successfully use small batches, e.g. $<1000$, for datasets such as \acrshort{mnist}. We always report exact \gls{d-cal} in the results. We evaluate the tightness of this bound and show that models ordered by the upper-bound are ordered in \gls{d-cal} the same way in \Cref{appsec:jensenslack}.

\subsection{Handling Censored Data}
\label{sec:handlingcensored}
In presence of right-censoring, failure times are censored more often than earlier times. So, applying the true \gls{cdf} to only uncensored failure times results in a non-uniform distribution skewed to smaller values in $[0,1]$. Censoring must be taken into account.

Let $x$ be a censored point with observed censoring time $u$ and unobserved failure time $\mbt$.  Recall that $\boldsymbol{\delta}=\indicator{\mbt<\mbc}$. In this case $\mbc=\mbu=u$ and $\boldsymbol{\delta}=0$.
Let $F_{\mbt} = F(\mbt \g x)$, $F_{\mbc}=F(\mbc \g x)$,
and $F_{\mbu} = F(\mbu \g x)$.
We first state the fact that,
under $\mbt \indep \mbc \g \mbx$, a datapoint observed to be censored at time $u$ has $F_{\mbt} \sim \text{Unif}(F_{u},1)$ for true \gls{cdf} $F$ (proof in  \Cref{appsec:cdfuniform}). This means that we can compute the probabilty that $\mbt$ falls in each bin $I=[a,b]$:
\begin{align}
\label{eq:expectedcount}
    \mathbb{P}(F_{\mbt} \in I \g 
    \delta=0, u, x) =
    \frac{(b-F_u)\indicator{F_u \in I}}{1-F_u} +
    \frac{(b-a)\indicator{F_u < a}}{1-F_u},
\end{align}
\cite{haider2020effective} make this observation and suggest a method for handling censoring points: they contribute $\mathbb{P}(F_{\mbt} \in I \g \delta=0,u,x)$ in place of the unobserved $\indicator{F_{\mbt} \in I}$:
    \begin{align}
    \label{eq:dcalcensored}
     \sum_{I \in \cI}
        \Big(
        \E_{\mbu,\boldsymbol{\delta},\mbx}
        \Big[
        \boldsymbol{\delta} \indicator{F_{\mbu} \in I}
        +
        (1-\boldsymbol{\delta})\mathbb{P}(F_{\mbt} \in I \g \boldsymbol{\delta},\mbu,\mbx) 
        \Big] 
        - |I|
        \Big)^2.
     \end{align}
This estimator does not change the expectation defining \gls{d-cal}, thereby preserving the property that \gls{d-cal} is $0$ for a calibrated model.  We soften \Cref{eq:expectedcount} with:
\begin{align*}
    \zeta_{\gamma,cens}(F_u; I)  := 
    \frac{(b-F_u)\sigma(\gamma(F_u-a)(b-F_u))}{(1-F_u)} +
    \frac{(b-a)\sigma(\gamma (a-F_u))}{(1-F_u)},
\end{align*}
where we have used a one-sided soft indicator for $\indicator{F_u < a}$ in the right-hand term.
We use $\zeta_{\gamma, cens}$ in place of $\zeta_{\gamma}$ for censored points in soft \gls{d-cal}. This gives the following estimator for soft \gls{d-cal} with censoring:
\begin{align}
    \label{eq:softdcalcensored}
     \sum_{I \in \cI}
        \Big(
        \E_{\mbu,\boldsymbol{\delta},\mbx}
        \Big[
        \boldsymbol{\delta} \zeta_\gamma(
        F_{\theta}(\mbu \g \mbx);I)
        +
        (1-\boldsymbol{\delta}) \zeta_{\gamma,cens}(
        F_{\theta}(\mbu \g \mbx);I
        )
        \Big] 
        - |I|
        \Big)^2.
     \end{align}
The upper-bound of \Cref{eq:softdcalcensored} and its corresponding gradient can be derived analogously to the uncensored case. We use these in ours experiments on censored data.

\section{Experiments \label{sec:experiments}}

We study how \gls{ours} allows the modeler to optimize for a specified balance between prediction and calibration.
We augment maximum likelihood estimation with
 \gls{ours} for various settings of coefficient $\lambda$, where $\lambda=0$ corresponds to vanilla maximum likelihood. Maximum likelihood for survival analysis is
described in  \Cref{appsec:appendixbackground} (\Cref{eq:likelihood}). For the log-normal experiments,
we also use \glsreset{s-crps}\gls{s-crps} \citep{avati87countdown} with \gls{ours}
since \gls{s-crps} enjoys a closed-form for log-normal. \Gls{s-crps} was developed to produce calibrated survival models but it optimizes neither a calibration measure nor a traditional likelihood. See \Cref{appsec:surv-crps} for a description
of \gls{s-crps}.

\paragraph{Models, Optimization, and Evaluation} We use log-normal, Weibull, Categorical and \gls{mtlr} models with various linear or deep parameterizations. For the discrete models, we optionally interpolate their \gls{cdf} (denoted in the tables by
\acrshort{ni} for not-interpolated and \acrshort{i} for interpolated). See \Cref{appsec:model} for general model descriptions. Experiment-specific model details may be found in  \Cref{appsec:experimentdetails}. We use $\gamma=10000$. We use $20$ \gls{d-cal} bins disjoint over $[0,1]$ for all experiments except for the cancer data, where we use $10$ bins as in \cite{haider2020effective}. 
 For all experiments, we measure the loss on a validation set at each training epoch to chose a model to report test set metrics with. We report the test set \gls{nll}, test set \gls{d-cal} and Harrell's Concordance Index  \citep{harrell1996multivariable}
(abbreviated \acrshort{conc}) on the test set for several settings of $\lambda$. We compute concordance using the Lifelines package \citep{davidson2017camdavidsonpilon}.
All reported results are an average of three seeds.

\paragraph{Data} We discuss differences in performance on simulated gamma data, semi-synthetic survival data where times are conditional on the \acrshort{mnist} classes, length of stay prediction in the \gls{mimic-iii} dataset \citep{johnson2016mimic},
and glioma brain cancer data from \gls{tcga}. Additional data details may be found in \Cref{appsec:datadetails}.

 \subsection{Experiment 1: Simulated Gamma Times with Log-Linear Mean}
 
 \paragraph{Data}
    We design a simulation study 
    to show that a conditional distribution may achieve good concordance and likelihood but will have poor \gls{d-cal}. After adding \gls{ours}, we are able to improve the exact \gls{d-cal}. We sample $\mbx \in \mathbb{R}^{32}$ from a multivariate normal with $\sigma^2=10.0$. We sample times $\mbt$ conditionally from a gamma with mean $\mbmu$
    that is log-linear in $\mbx$ and constant variance 1e-3.
The censoring times $\mbc$ are drawn like the event times, except with a different coefficient for the log-linear function.  We experiment with censored and uncensored simulations,
where we discard $\mbc$ and always observe $\mbt$ for uncensored.
We sample a train/validation/test sets with 100k/50k/50k datapoints, respectively. 

\paragraph{Results} Due to high variance in $\mbx$ and low conditional variance, this simulation has low noise. With large, clean data, this experiment validates the basic method on continuous
and discrete models in the presence of censoring. \Cref{tab:gammacensored} demonstrates how increasing $\lambda$ gracefully balances \gls{d-cal} with \gls{nll} and concordance for different models and objectives: log-normal trained via \gls{nll} and with \gls{s-crps}, and the categorical model trained via~\gls{nll}, without \gls{cdf} interpolation. 
For results on more models and choices of $\lambda$ see  \Cref{tab:gammauncensoredfull} for uncensored results
and \Cref{tab:gammacensoredfull} for censored in \Cref{appsec:fullresults}.

\begin{table}[ht]
\centering
\caption{\label{tab:gammacensored} Gamma simulation, censored}
% \begin{tabular}{llSSSSSSSS}
\begin{tabular}{llllllllll}
\toprule 
& $\lambda$ &0&1&10&100&500&1000\\ 
\midrule 
Log-Norm & \acrshort{nll} &-0.059&-0.049&0.004&0.138&0.191&0.215\\
\gls{nll} & \acrshort{d-cal-short}&0.029&0.020&0.005&2e-4&6e-5&7e-5\\
&\acrshort{conc}&0.981&0.969&0.942&0.916&0.914&0.897\\ \midrule 
Log-Norm & \acrshort{nll}&0.038&0.084&0.143&0.201&0.343&0.436\\
\gls{s-crps} & \acrshort{d-cal-short}&0.017&0.007&0.001&1e-4&5e-5&8e-5\\
& \acrshort{conc}&0.982&0.978&0.963&0.950&0.850&0.855\\
\midrule
Cat-\acrshort{ni} & \acrshort{nll}&0.797&0.799&0.822&1.149&1.665&1.920\\
 & \acrshort{d-cal-short}
&0.009&0.006&0.002&2e-4&6e-5&6e-5\\
& \acrshort{conc}
&0.987&0.987&0.987&0.976&0.922&0.861\\\bottomrule
\end{tabular}
\end{table}

\subsection{Experiment 2: Semi-Synthetic Experiment: Survival MNIST}

\paragraph{Data} Following \cite{polsterl2019survivalmnist}, we simulate a survival dataset conditionally on the \acrshort{mnist} dataset \citep{lecun2010mnist}. Each \acrshort{mnist} label gets a deterministic risk score, with labels loosely grouped together by risk groups (\Cref{tab:risks} in \Cref{appsec:mnistdetails}). Datapoint image $\mbx_i$ with label $\mby_i$ has time $\mbt_i$ drawn from a Gamma with mean equal to $\text{risk}(y_i)$
and constant variance 1e-3. Therefore
$\mbt_i \indep \mbx_i \g \mby_i$
and times for datapoints that share an \acrshort{mnist} class are identically drawn.
We draw censoring times $\mbc$ uniformly between the minimum failure time and the $90^{th}$ percentile time, which resulted in about 50\% censoring. We use PyTorch's \acrshort{mnist} with test split into validation/test. The model does not see the \acrshort{mnist} class and learns a distribution over times given pixels $\mbx_i$.
We experiment with censored and uncensored simulations,
where we discard $\mbc$ and always observe $\mbt$ for uncensored.

\paragraph{Results}
This semi-synthetic experiment tests the ability to
tune calibration in presence of
a high-dimensional conditioning set (\acrshort{mnist} images) and through a typical convolutional architecture. \Cref{tab:mnistcensored} demonstrates that the deep log-normal models started
off miscalibrated relative to the categorical model for $\lambda=0$ and that all models were able to significantly improve in calibration.
See
\Cref{tab:mnistuncensoredfull} and \Cref{tab:mnistcensoredfull} for more uncensored and censored survival-\acrshort{mnist} results.

\begin{table}[ht]
\centering
\caption{\label{tab:mnistcensored} Survival-\acrshort{mnist}, censored}
 \begin{tabular}{lllllllll}
\toprule
& $\lambda$&0&1&10&100&500&1000\\
\midrule
Log-Norm & \acrshort{nll} &4.337&4.377&4.483&4.682&4.914&5.151\\
\gls{nll} & \acrshort{d-cal-short}
&0.392&0.074&0.020&0.005&0.005&0.007\\
& \acrshort{conc}
&0.902&0.873&0.794&0.696&0.628&0.573\\ \midrule 
Log-Norm & \acrshort{nll}
&4.950&4.929&4.859&4.749&4.786&4.877\\
\gls{s-crps} & \acrshort{d-cal-short}
 &0.215&0.122&0.051&0.010&0.002&9e-4\\
& \acrshort{conc}
&0.891&0.881&0.874&0.868&0.839&0.815\\ \midrule 
Cat-\acrshort{ni} & \acrshort{nll}
&1.733&1.734&1.765&1.861&2.074&3.030\\
& \acrshort{d-cal-short}
&0.018&0.014&0.004&5e-4&5e-4&4e-4\\
& \acrshort{conc}
&0.945&0.945&0.927&0.919&0.862&0.713\\\bottomrule
\end{tabular}
\end{table}

\subsection{Experiment 3: Length of Stay Prediction in MIMIC-III}  

\paragraph{Data} We predict the length of stay (in number of hours) in the ICU, using data from the \acrshort{mimic-iii}
dataset. Such predictions are important
both for individual risk predictions and prognoses and for hospital-wide resource management. We follow the preprocessing in \cite{harutyunyan2017multitask}, a popular \acrshort{mimic-iii} benchmarking paper and repository \footnote{https://github.com/YerevaNN/mimic3-benchmarks}. The covariates are a time series of $17$ physiological variables~(\Cref{tab:variables} in
\Cref{appsec:mimicdetails}) including respiratory rate and glascow coma scale information.
There is no censoring in this task. 
We skip imputation and instead use missingness masks as features.
There are $2,925,434$ and $525,912$ instances in the training and test sets. We split the training set in half for train and validation.

\paragraph{Results} 
\cite{harutyunyan2017multitask} discuss the difficult of this task
when predicting fine-grained lengths-of-stay, as opposed to simpler classification
tasks like more/less one week stay. The true conditionals are high in entropy given the chosen covariates \Cref{tab:newmimic} 
demonstrates this difficulty, as can be seen in the concordances. We report the categorical model with and without \gls{cdf} interpolation and the log-normal trained with \gls{s-crps}. \Gls{nll} for the log-normal is not reported because \gls{s-crps} does not optimize \gls{nll} and did poorly on this metric. The log-normal trained with \gls{nll} was not able to fit this task on any of the three metrics. All three models reported are able to reduce \gls{d-cal}.
Results for all models and more choices of $\lambda$ may be found in  \Cref{tab:newmimicfull}.
The categorical models with and without \gls{cdf} interpolation match in concordance for $\lambda=0$ and $\lambda=1000$. However, the interpolated model achieves better \gls{d-cal}.
This may be due to the lower-bound $\ell>0$ on a discrete model's \gls{d-cal} (\Cref{appsec:model}). 

\begin{table}[ht]
\centering
\caption{\label{tab:newmimic} \acrshort{mimic-iii} length of stay}
 \begin{tabular}{lllllllll}
\toprule
& $\lambda$&0&1&10&100&500&1000\\\midrule
Log-Norm   & \acrshort{d-cal-short}
&0.859&0.639&0.155&0.046&0.009&0.005\\
\gls{s-crps} & \acrshort{conc}
&0.625&0.639&0.575&0.555&0.528&0.506\\\midrule 
Cat-\acrshort{ni} &  Test  \gls{nll}
&3.142&3.177&3.167&3.088&3.448&3.665\\
 & \acrshort{d-cal-short}
&0.002&0.002&0.001&2e-4&1e-4&1e-4\\
& \acrshort{conc}
&0.702&0.700&0.699&0.690&0.642&0.627\\\midrule 
Cat-\acrshort{i} & \acrshort{nll}
&3.142&3.075&3.073&3.073&3.364&3.708\\
 & \acrshort{d-cal-short}
&4e-4&2e-4 &2e-4 &1e-4&5e-5&4e-5\\
& \acrshort{conc}
&0.702&0.702 &0.702&0.695&0.638&0.627\\\bottomrule
\end{tabular}
\end{table}

\subsection{Experiment 4: Glioma data from The Cancer Genome Atlas}

We use the glioma (a type of brain cancer) dataset \footnote{https://www.cancer.gov/about-nci/organization/ccg/research/structural-genomics/tcga/studied-cancers/glioma} collected as part of the \gls{tcga} program and studied in~\citep{cancer2015comprehensive}.
We focus on predicting time until death from the clinical data, which includes tumor tissue location, time of pathological diagnosis,
Karnofsky performance score, radiation therapy, demographic information,
and more. Censoring means they did not pass away. The train/validation/test sets are made of 552/276/277 datapoints respectively, of which 235/129/126 are censored, respectively.

\paragraph{Results}
For this task, we study the Weibull \gls{aft} model, 
reduce the deep log-normal model from three to two hidden layers,
and study a linear \gls{mtlr} model (with \gls{cdf} interpolation) in place of the deep categorical due to the small data size. \Gls{mtlr} is more constrained than linear categorical due to shared parameters. \Cref{tab:gbmlgg} demonstrates these three models' ability to improve \gls{d-cal}.
\gls{mtlr} is able to fit well and does not give up
much concordance.
Results for all models and more choices of $\lambda$ may be found in \Cref{tab:gbmlggfull}.

\begin{table}[ht]
\centering
\caption{\label{tab:gbmlgg} The Cancer Genome Atlas, glioma}
%  \begin{tabular}{ccccccccccc}
%   \begin{tabular}{llSSSSSSSSS}
\begin{tabular}{llllllllll}
\toprule
&$\lambda$&0&1 & 10 &100&500&1000\\\midrule
%  &  & & Log-Normal &NLL & & & & \\ \midrule 
Log-Norm & 
 \gls{nll}
&14.187&6.585&4.639&4.181&4.403&4.510\\
\gls{nll} & \acrshort{d-cal-short}
&0.059 &0.024 &0.010 &0.003&0.002&0.004\\
& \acrshort{conc}
&0.657&0.632 &0.703&0.805&0.474&0.387\\ \midrule 
%  &  & & Weibull &NLL & & & & \\ \midrule 
 Weibull & \gls{nll}
&4.436&4.390 &4.292&4.498&4.475&4.528\\
& \acrshort{d-cal-short}
&0.035 &0.028 &0.009&0.003&0.004&0.007\\
& \acrshort{conc}
&0.788&0.785 &0.777&0.702&0.608&0.575\\\midrule 
%  &  & & mtlr &NLL & & & & \\ \midrule 
\gls{mtlr}-\acrshort{ni} & \gls{nll}
&1.624&1.620&1.636&1.658&1.748&1.758\\
& \acrshort{d-cal-short}
&0.009&0.007&0.005&0.003&0.002&0.002\\
& \acrshort{conc}
&0.828&0.829&0.824&0.818&0.788&0.763\\\bottomrule
\end{tabular}
\end{table}

\section{Related Work}

\paragraph{Deep Survival Analysis} Recent approaches to survival analysis 
parameterize the failure distribution as a deep neural network function of the \citep{ranganath2016deep, alaa2017deep, katzman2018deepsurv}.
\cite{miscouridou2018deep} and \cite{lee2018deephit} use a discrete categorical distribution over times interpreted ordinally, which can approximate any smooth density
with sufficient data.
The categorical approach has also been used when the conditional is parameterized by a recurrent neural network of sequential covariates \citep{giunchiglia2018rnn,ren2019deep}.
\cite{miscouridou2018deep} extend deep survival analysis to deal with missingness in $\mbx$.

\paragraph{Post-training calibration methods}
Practitioners have used two calibration methods for binary classifiers, which modify model predictions maximize likelihood on a held-out dataset. Platt scaling \citep{platt1999probabilistic} works by using a scalar logistic regression built on top of predicted probabilities.
Isotonic regression \citep{zadrozny2002transforming} uses a nonparametric piecewise linear transformation instead of the logistic regression.
These methods do not reveal an explicit balance between prediction quality and calibration during model training.
\Gls{ours} allows practitioners to explore this balance while searching in the full model space.

\paragraph{Objectives}
When an unbounded loss function (e.g. \gls{nll}) is used and the gradients are a function of $x$, the model may put undue focus on explaining a given outlier $x^\star$, worsening calibration during training. For this reason,
robust objectives have been explored.
\citet{avati87countdown} consider \gls{crps} \citep{matheson1976scoring}, a robust proper scoring rule for continuous outcomes, and adapt it to \gls{s-crps} for survival analysis by accounting for censoring.
However, \gls{s-crps} does not provide a clear way to balance predictive power and calibration. \cite{kumar2018trainable} develop a trainable kernel-based calibration measure for binary classification but they do not discuss an optimizable calibration metric for survival analysis.

\paragraph{Brier Score}
The Brier Score \citep{brier1951verification}
decomposes into a calibration metric (numerator of Hosmer-Lemeshow) and a discrimination term encouraging patients 
with the same failure status at $t^\star$ to have the same failure probability at $t^\star$. To capture entire distributions over time, the Integrated Brier Score is used. The Inverse Probability of Censoring Weighting Brier Score \citep{graf1999assessment} handles censoring but requires estimation of the censoring distribution,  a whole survival analysis problem (with censoring due to the failures) on its own \citep{gerds2006consistent,kvamme2019brier}. \Gls{ours} can balance discrimination and calibration without estimation of the censoring distribution.

\section{Discussion} 

Model calibration is an important consideration in many clinical problems, especially when treatment decisions require risk estimates across all times in the future.
We tackle the problem of building models that are calibrated over individual failure distributions. 
To this end, we provide a new technique that explicitly targets calibration during model training.
We achieve this by constructing a differentiable approximation of \gls{d-cal}, and using it as an add-on objective to maximum likelihood and \gls{s-crps}.
As we show in our experiments, \gls{ours} allows for explicit and direct control of calibration on both simulated and real data.
Further, we showed that searching over the \gls{ours} $\lambda$ parameter can strike the practitioner-specified balance between predictive power and calibration. 

\paragraph{Marginal versus Conditional Calibration} \gls{d-cal} is $0$ for the true conditional and marginal distributions of failure times. This is because
\gls{d-cal} measures marginal calibration, i.e. $\mbx$ is integrated out. Conditional calibration is the stronger condition that $F_\theta(t \g x)$ is calibrated for all $x$. This is in general infeasible even to measure (let alone optimize)
\citep{vovk2005algorithmic,pepe2013methods,barber2019limits}
without strong assumptions since for continuous $x$ we usually observe just one sample. 
However, among the distributions that have $0$ \gls{d-cal},
the true conditional distribution has the smallest \gls{nll}.
Therefore, \xcaled{} objectives with proper scoring rules (like \gls{nll}) have an optimum only for the true conditional model in the limit of infinite data.

\paragraph{D-Calibration and Censoring}
\Cref{eq:dcalcensored} in  \Cref{sec:handlingcensored} provides a censored version of \gls{d-cal} that is $0$ for a calibrated model, like the original \gls{d-cal} (\Cref{eq:dcal}). However, this censored calibration measure
is not equal to \gls{d-cal} in general for miscalibrated models. For a distribution $F_\theta$ with non-zero \gls{d-cal}, for any censoring distribution $G$, estimates of the censored version will assess $F_\theta$ to be more uniform than if exact \gls{d-cal} were able to be computed using all true observed failure times. This happens especially in the case of heavy and early censoring because a lot of uniform weight is assigned \citep{haider2020effective,avati87countdown}. This means that the censored objective can be close to $0$ for a miscalibrated model on a highly censored dataset. 

An alternative strategy that avoids this issue is to use inverse weighting methods (e.g. Inverse Propensity Estimator of outcome under treatment \citep{horvitz1952generalization},
Inverse Probability of Censoring-Weighted Brier Score
\citep{graf1999assessment,gerds2006consistent}
and Inverse Probability of Censoring-Weighted binary calibration for survival analysis
\citep{yadlowsky2019calibration}). Inverse weighting would preserve
the expectation that defines \gls{d-cal} for any censoring distribution. One option is to adjust with $p(\mbc \mid \mbx)$. This requires $\mbc \perp \mbt \mid \mbx $ and solving an additional censored survival problem $p(\mbc \mid \mbx)$.  Nevertheless, if a censoring estimate is provided, the methodology in this work could then be applied to an inverse-weighted \gls{d-cal}.
There is then a trade-off between the censored estimator proposed by \cite{haider2020effective} that we use (no modeling $G$) and inverse-weighted estimators (which preserve \gls{d-cal} for miscalibrated models).

\section*{Broader Impact}

In this paper, we study calibration of survival analysis models and suggest an objective for improving calibration during model training. Since calibration means that modeled probabilities correspond to the actual observed risk of an event, practitioners may feel more confident about using model outputs directly
for decision making e.g. to decide how many emergency room staff members qualified for performing a given procedure should be present tomorrow given all current ER patients. But if the distribution of event times in these patients differs from validation data, because say the population has different demographics, calibration should not provide the practitioner with more confidence to directly use such model outputs.

\section*{Acknowledgments}
This work was supported by:
\begin{itemize}
    \item NIH/NHLBI Award R01HL148248 
    \item NSF Award 1922658 NRT-HDR: FUTURE Foundations, Translation, and Responsibility for Data Science.
    \item NSF Award 1514422 TWC: Medium: Scaling proof-based verifiable computation
    \item NSF Award 1815633 SHF
\end{itemize}

We thank Humza Haider for sharing the original D-calibration experimental data, \citet{avati87countdown} for publishing their code and the Cancer Genome Atlas Research Network for making the glioma data public. We thank all the reviewers for thoughtful feedback.

\newpage

\appendix

\section{\label{appsec:appendixbackground} Background on Survival Analysis and Related Work}

Survival analysis models the probability distribution of a time-until-event. The event is often called a failure time. For example, we may model time until onset of coronary heart disease given a patient's current health status \citep{wilson1998prediction,d2008general}.

Survival analysis differs from standard probabilistic regression problems in that data may be censored. For example, 
a patient may leave a study before developing the studied condition, or may not develop the condition before the study ends.
In these cases, the time that a patient leaves or the study ends is called the censoring time. These are cases of right-censoring, 
where it is only known that the failure time is greater than the observed censoring time.

We review key definitions in survival analysis. See \cite{george2014survival} for a review. For textbooks, see \cite{andersen2012statistical}, \cite{kalbfleisch2002}, and \cite{lawless2011}. 

\subsection{Notation}

Let $\mbt$ be a continuous random variable denoting the failure time with \gls{cdf} $F$
and density $f$. The survival function $\overline{F}$ is defined as 1 minus the \gls{cdf}: $\overline{F}=1-F$. Censoring times are considered random variables $\mbc$ with \gls{cdf} $G$, survival function $\overline{G}$, and density $g$. In general these distributions may be conditional on covariates $\mbx$.

For datapoints $i$, let $\mbt_i$ be failure times and $\mbc_i$ be censoring times. Let us focus on right-censoring where $\mbu_i=\min(\mbt_i,\mbc_i)$, $\boldsymbol{\delta}_i=\indicator{\mbt_i < \mbc_i}$ and the observed data consists of $(x_i,u_i,\delta_i)$. In general we cannot throw away censored points, since $p(\vt \g x, \vt < c) \neq p(\vt \g x)$ and we would therefore biasedly estimate the failure distribution $F$.

\subsection{Assumptions About Censoring}

It may seem that we need to model $\mbc$ to estimate the parameters of $f$, but under certain assumptions, we can write the likelihood (with respect to $f$'s parameters) for a dataset with censoring without estimating the censoring distribution. In this work, we assume:
\begin{assumption*}
Censoring-at-random. $\mbt$ is distributed marginally or conditionally on $\mbx$. $\mbc$ is either a constant, distributed marginally, or distributed conditionally on $\mbx$. In any case, it must hold that $\mbt \indep \mbc \mid \mbx$.
\end{assumption*}

\begin{assumption*}
Non-informative Censoring.
The censoring time $\mbc$'s distribution parameters $\theta_c$
are distinct from parameters
$\theta_t$ of $\mbt$'s distribution.
\end{assumption*}

\subsection{Likelihoods} Under the two censoring assumptions, the log-likelihood can be derived to be
\begin{align}
\label{eq:likelihood}
L(\theta_t) = \sum_i \delta_i\log f_{\theta_t}(t_i \mid x_i)
+ (1 - \delta_i )\log \overline{F}_{\theta_t}(t_i|x_i)
\end{align}
and can be maximized to learn parameters $\theta_t$ of $f$ 
without an estimate of $G$.  This can be interpreted as follows: an uncensored individual has $\delta_i=1$, meaning $u_i=t_i$. This point contributes through the failure density $f(u_i)=f(t_i)$, as in standard regression likelihoods. Censored points contribute through failure survival function $\overline{F}=1-F$ because there failure time is known to be greater than $u_i$. Full discussions of survival likelihoods can be found in 
\cite{kalbfleisch2002,lawless2011,andersen2012statistical}. 

\subsection{Testing Calibration}
Classical goodness-of-fit tests
\citep{lemeshow1982review,gronnesby1996method,d2003evaluation} and their recent modifications 
\citep{demler2015tests} assess calibration of survival analysis models
for a particular time of interest $\vt^*$. These take the following steps:

\begin{enumerate}
    \item  pick a time $t^\star$ at which to measure calibration 
    
    \item evaluate model probability $p_i = p_\theta(\mbt < t^\star \g \mbx_i)$
    of failing by time $t^\star$

\item sort $p_i$ into $K$ groups $g_k$ defined by quantiles (e.g. $K=2$ corresponds to partitioning the data into a low-risk group and high-risk group)

\item compute the \textit{observed} $\#$
of events using e.g. $(1-\acrshort{km}_k[\vt^*]) |g_k|$ where $\acrshort{km}_k$ the Kaplan-Meier estimate \citep{kaplan1958nonparametric} of the survival function just on data in $g_k$'s

\item compute the \textit{expected} $\#$,
$E_k= \sum_{i \in g_k} p_i$

\item let $\overline{p}_k=\frac{1}{|g_k|} \sum_{i \in g_k} p_i$

\item 
$\sum_k  \frac{(O_k-E_k)^2}{|g_k|\overline{p}_k (1-\overline{p}_k)}$
gives a $\chi^2$ test statistic

\item small p-value $\rightarrow$ model not calibrated

\end{enumerate}

\cite{demler2015tests} review these
tests and propose some modifications when there are not enough individuals assigned to each bin. These tests are limited in two ways: they answer calibration in a  rigid yes/no fashion with hypothesis testing, and it is not clear how to combine calibration assessments over the entire range of possible time predictions.

\subsection{Calibration Slope}\label{appsec:slope}

\paragraph{Calibration Slope} 
Recent publications in machine learning
\citep{avati87countdown}
and in medicine
\citep{besseling2017selection}
use the \textit{calibration slope}
to evaluate calibration \citep{stevens2020validation}.
First, a calibration curve is computed by plotting,
for each quantile $\rho\in[0,1]$,
the fraction of observed samples with a failure time smaller than
that quantile's time $\vt(\rho)=F^{-1}_\theta(\rho\g x)$.
Then, report the slope of the best-fit line to this curve.
When a model is well-calibrated, the true and predicted densities are close and the best fit line has slope $1.0$.
However, slope can be $1.0$ (with intercept $0.0$) even when the model is not well-calibrated. 

\begin{figure}[t]
  \centering 
  \includegraphics[width=0.7\textwidth]{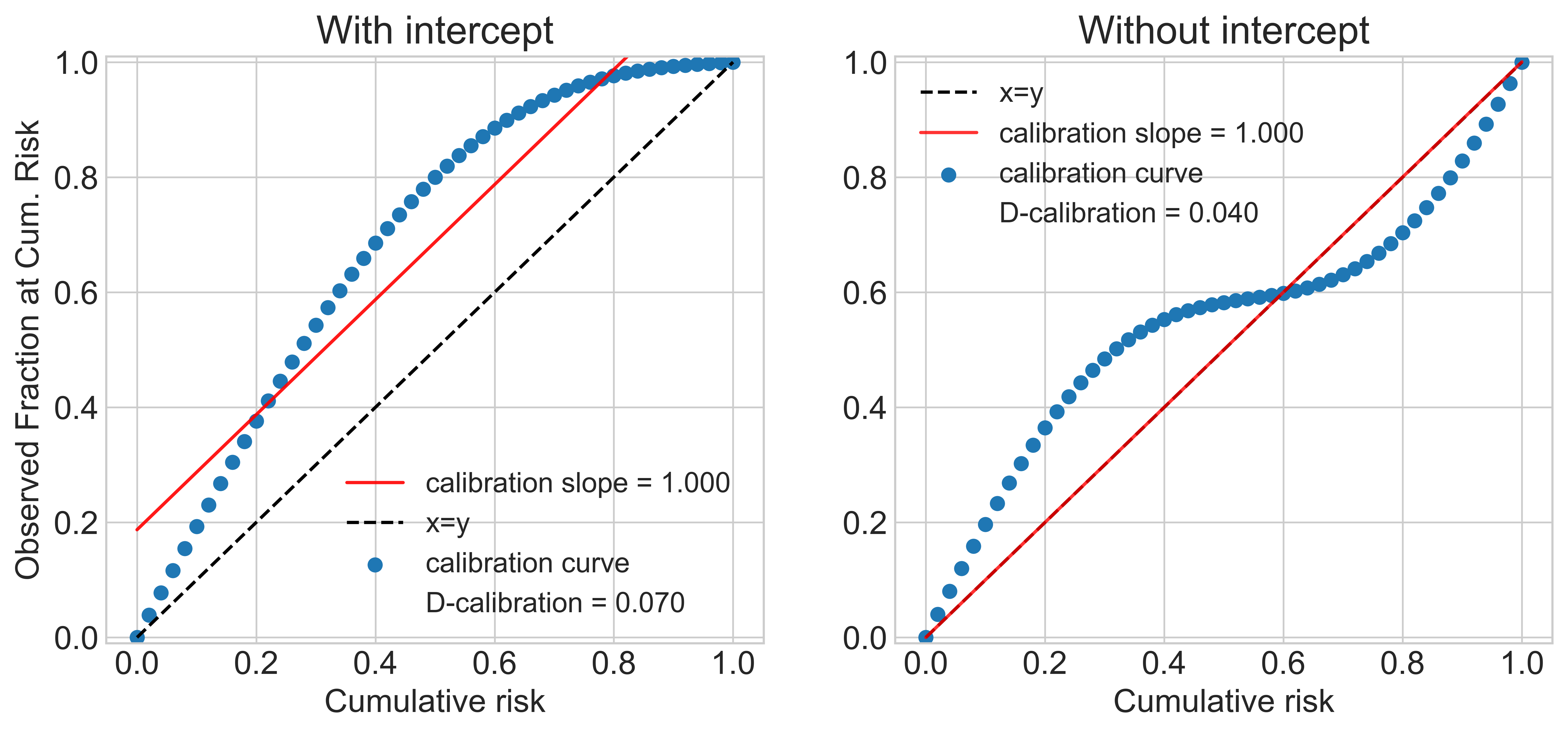} 
  \caption{Sub-optimal calibration curves that result in optimal calibration slope.}
  \label{fig:slope-is-bad} 
\end{figure}

Here, we construct two possible calibration curves that cannot result from well-calibrated models. However, the resulting calibration slope is close to $1.0$.
\cite{avati87countdown} use a line
of best fit with non-zero intercept.
We plot hypothetical calibration curves in~\Cref{fig:slope-is-bad} such that the corresponding best fit line has slope $1.0$, with and without intercept terms.
\cite{stevens2020validation} make a related observation about calibration slope: a near-zero intercept of the line of best fit, or other evidence of calibration, should always be reported alongside near-1 slope when claiming a model is calibrated. However, we demonstrate here that even slope $1$ and intercept $0$ can result from
poorly calibrated models. The interested reader should see \cite{stevens2020validation} for an assessment of recent publications in medicine that report only slope and for the history of slope-only as a ``measure
of spread"  \citep{cox1958two}.

\section{Survival CRPS \label{appsec:surv-crps}}
\gls{s-crps} is proposed by \citet{avati87countdown}:
\[
\mathcal{S}_{\text{CRPS}}(\hat{F},(y,c)) = \int_0^y \hat{F}(z)^2 dz + (1-c) \int_y^\infty (1-\hat{F}(z))^2 dz,
\]
where $y$ is the event time, $c$ is an indicator for censorship and $\hat{F}$ is the \gls{cdf} from the model.
See \citet{avati87countdown} Appendix B for a detailed derivation of \gls{s-crps} objective for a log-normal model.

\section{CDF of Survival Time is Uniform for Censored Patient}
\label{appsec:cdfuniform}

Consider the data distribution $P(\mbt, \mbc \g x)$
and using the conditional $P(\mbt \g x)$ of this distribution to evaluate \gls{d-cal} on this data.
For a point that is censored at time $c$, $P(\mbt \g x)$
would simply condition on the event $\mbt > c$ for constant $c$, yielding $P(\mbt \g \mbt > c,x)$. However, the true failure distribution for such a point is $P(\mbt \g \mbt > c,\mbc=c,x)$.
Under censoring-at-random,
\begin{align}
\label{eq:noninformative}
\mbt \indep \mbc \g \mbx
\implies 
P(\mbt \g \mbt>c, x) 
= P(\mbt \g \mbt > c,\mbc=c,x).
\end{align}
Let $F$ be the failure \gls{cdf}. 
Let $p_{t}$ be the density of $\mbt \g x$. Apply transformation $\mbz = F(\mbt | x)$. To compute $\mbz$'s density, we need:
\begin{align*}
    \frac{d}{dz} F^{-1}(z|x) = \frac{1}{p_{t}(F^{-1}(z \mid x))} = \frac{1}{p_t(t)}.
\end{align*}

Applying change of variable to compute $\mbz$'s density:
\begin{align*}
    p_t(F^{-1}(z|x))  \frac{d}{dz}F^{-1}(z|x) = p_t(t)  \frac {1}{p_t(t)} = 1
\end{align*}
Therefore, $\mbz$ is uniform distributed over [0,1]. So conditioning on set
$(\mbt > c, x) = (\mbz > F(c|x), x)$ gives the result:
\begin{align*}
\label{eq:unifcensor}
    \mbz \g (\mbt > c,x)  \sim \text{Unif}(F(c \g x),1).
\end{align*}

The \gls{cdf} value of the unobserved time for a censored datapoint is uniform above the failure \gls{cdf} applied to the censoring time. \cite{haider2020effective} (Appendix B) give an alternate proof in terms of expected counts.

\section{Extra Data Details \label{appsec:datadetails}}

\subsection{Data Details for Simulation Study \label{appsec:gammadatadetails}}

For the gamma simulation, we draw $\mbx$
from a $D=32$ multivariate Normal
with $\mathbf{0}$ mean and
diagonal covariance with $\sigma^2=10.0$.
We draw failure times $\mbt$ conditionally on $\mbx$ from a gamma 
distribution with mean $\mbmu$ log-linear in $\mbx$. The weights
of the linear function are drawn uniformly.
The gamma distribution has constant variance 1e-3.
This is achieved by setting
$\alpha = \mbmu_i^2 / 1e\text{-}3$ and
$\beta = \mbmu_i / 1e\text{-}3$.
\begin{align*}
\mbx_i \sim \cN(0, \sigma^2  \mathbf{I}),
\quad
\mathbf{w}_d \sim  \texttt{Unif}(-0.1,0.1) , \quad 
 \mbmu_i = \exp[\mathbf{w}^\top \mbx_i], \quad
    \mbt_i \sim \mathtt{Gamma}(\alpha,\beta).
\end{align*}
Censoring times are drawn like failure times but with
a different set of weights for the linear function.
This means $\mbt \indep \mbc \g \mbx$.

\subsection{Data Details for MNIST \label{appsec:mnistdetails}}

As described in the main text, we follow \cite{polsterl2019survivalmnist}
to  simulate a survival dataset conditionally on the \acrshort{mnist} dataset \citep{lecun2010mnist}. 
Each \acrshort{mnist} label gets a deterministic risk score, with labels loosely grouped together by risk groups.
See \Cref{tab:risks} for an example of the risk groups and risk scores for the \acrshort{mnist} classes.

Datapoint image $\mbx_i$ with label $\mby_i$ has time $\mbt_i$ drawn from a Gamma whose mean is the risk score and whose variance is constant 1e-3. Therefore
$\mbt_i$ is independent of $\mbx_i$ given $\mby_i$
and times for datapoints that share an \acrshort{mnist} class are identically drawn.
\begin{align*}
    \boldsymbol{\mu}_i = \text{risk}(\mby_i) \quad v = 1e\text{-}3 \quad 
 \alpha = \mbmu_i^2 / v, \quad 
    \beta = \mbmu_i / v, \quad
    \mbt_i \sim \mathtt{Gamma}(\alpha,\beta)
\end{align*}

For each split of the data (e.g. training set),
we draw censoring times uniformly between the  minimum failure time in that split and the $90^{th}$ percentile time, which, due to the particular failure distributions, resulted in about 50\% censoring.

\begin{table}[t]
    \centering
        \caption{\label{tab:risks} Risk scores for digit classes.}
    \begin{tabular}{ccccccccccc}
\toprule
    Digit  & 0 & 1 & 2 & 3 & 4 & 5 & 6 & 7 & 8 & 9 \\
    \midrule
    Risk Group  & most & least & lower & lower & lower & higher & least & most & least & most \\
    Risk Score  & 11.25 & 2.25 & 5.25 & 5.0 & 4.75 & 8.0 & 2.0 & 11.0 & 1.75 & 10.75 
    \\ \bottomrule \\
    \end{tabular}
\end{table}

\subsection{Data Details for MIMIC-III \label{appsec:mimicdetails}}
We show the 17 physiological variables we use in \Cref{tab:variables}. The table is reproduced from
\cite{harutyunyan2017multitask}. This dataset differs from other \acrshort{mimic-iii} length of stay datasets
because one stay in the ICU of a single patient produces many datapoints: 
remaining time at each hour after admission.
%The data processed by \cite{harutyunyan2017multitask} 
After excluding ICU transfers and patients under 18, 
there are $2,925,434$ and $525,912$ instances in the training and test sets. We split the training set in half for train and validation.

\begin{table}[t]
    \centering
        \caption{The 17 selected clinical variables. The second column shows the source table(s) of a variable from \acrshort{mimic-iii} database. The third column lists the ``normal'' values used in the imputation step. Table reproduced from \cite{harutyunyan2017multitask}.}
    \begin{tabular}{lll}
    \toprule
    Variable  table & Impute value & Modeled as \\
    \midrule
        Capillary refill rate  & 0.0 & categorical \\
        Diastolic blood pressure  & 59.0 & continuous \\
        Fraction inspired oxygen  & 0.21 & continuous \\
        Glascow coma scale eye opening  & 4 spontaneously & categorical \\
        Glascow coma scale motor response  & 6 obeys commands & categorical \\
        Glascow coma scale total  & 15 & categorical \\
        Glascow coma scale verbal response  & 5 oriented & categorical\\
        Glucose  & 128.0 & continuous \\
        Heart Rate  & 86 & continuous \\
        Height  & 170.0 & continuous \\
        Mean blood pressure  & 77.0 & continuous\\
        Oxygen saturation  & 98.0 & continuous\\
        Respiratory rate  & 19 & continuous\\
        Systolic blood pressure  & 118.0  & continuous\\
        Temperature  & 36.6 & continuous\\
        Weight  & 81.0 & continuous\\
        pH & 7.4 & continuous\\
    \bottomrule \\
    \end{tabular}
    \label{tab:variables}
\end{table}

\subsection{Data Details for The Cancer Genome Atlas Glioma 
 Data \label{appsec:gliomadetails}}
 
We use the glioma (a type of brain cancer) data\footnote{\href{https://www.cancer.gov/about-nci/organization/ccg/research/structural-genomics/tcga/studied-cancers/glioma}{https://www.cancer.gov/about-nci/organization/ccg/research/structural-genomics/tcga/studied-cancers/glioma}} collected as part of the \gls{tcga} program and studied in~\citep{cancer2015comprehensive}.
\Gls{tcga}  comprises clinical data and molecular from 11,000 patients being treated for a diverse set of cancer types. We focus on predicting time until death from the clinical data, which includes:
\begin{itemize}
    \item tumor tissue site
    \item time of initial pathologic diagnosis
    \item radiation therapy
    \item  Karnofsky performance score
    \item histological type
    \item demographic information
\end{itemize}
Censoring means they did not pass away. The train/validation/test sets are made of 552/276/277 datapoints respectively, of which 235/129/126 are censored, respectively. 

To download this data, use the \href{http://firebrowse.org/}{firebrowse}.
tool, select the Glioma (GBMLGG) cohort, and then click the blue clinical
features bar on the right hand side. Select the ``Clinical Pick Tier 1" file.

We standardized the features and then clamped their maximum absolute
value at $5.0$. This is in part because we were working with the Weibull \gls{aft} model, which is very sensitive to large variance in covariates.

\section{\label{appsec:model}Model Descriptions}
We describe the models we use in the experiments. 
For all models, the parameterization as a function of $\mbx$ varies
in complexity (e.g. linear or deep) depending on task. 

\paragraph{Log-normal model} When $\log T$ is Normal with mean $\mu$ and variance $\sigma^2$, we say
that $T$ is log-normal with location $\mu$ and scale $\sigma$. We parameterize $\mu$ and $\sigma$ as functions of $\mbx$ (small \texttt{ReLU} networks with $1$ to $3$ hidden layers, depending on experiment).

\paragraph{Weibull Model} The Weibull 
\glsreset{aft} \gls{aft} model sets $\log T = \beta_0 + \beta^\top X + \sigma W$ where $\sigma$ is a scale parameter and $W$ is Gumbel. It follows that
$T \sim \text{Weibull}(\lambda, k)$
with scale $\lambda=\exp[\beta^\top X]$
and concentration $k=\sigma^{-1}$ \citep{liu2018using}. We constrain $k \in (1,2)$.

\paragraph{Interpolation for Discrete Models}
The next two models predict for a finite set of times and therefore have a discontiuous \gls{cdf}.
These models have a lower-bound $\ell >0$ on \gls{d-cal} because the \gls{cdf} values will not be $\text{Unif}(0,1)$ distributed. However, $\ell$ decreases to $0$ as the number of discrete times increases. For any fixed number of times, minimizing \gls{d-cal} will still improve calibration, which we observe in our experiments. 

We optionally use linear interpolation to calculate the \gls{cdf}. Suppose a time $t$ falls into bin $k$ which covers time interval $(t_a,t_b)$. If we do not use interpolation, then the \gls{cdf} value $P(T\leq t)$ we calculate is the sum of the probabilities of bins whose indices are smaller than or equal to  $k$. If we use linear interpolation, we replace the probability of bin $k$, $P(k)$, in the summation by:
\begin{align*}
    \frac{t-t_a}{t_b-t_a} P(k)
\end{align*}

\paragraph{Categorical Model}
We parameterize a categorical distribution
over discrete times by using a neural network function of $\mbx$ with a size $B$ output. Interpreted ordinally, this can approximate continuous survival distributions as $B \to \infty$ \citep{lee2018deephit, miscouridou2018deep}. The time for each bin is set to training data percentiles so that each next bin captures the range of times for the next $(100/B)^{th}$ percentile of training data, using only uncensored times.

\paragraph{Multi-Task Logistic Regression (mtlr)} \gls{mtlr} differs from the Categorical Model because there is some relationship between the probability of the bins. Assume we have $K$ bins. In the linear case \citet{yu2011learning}, suppose our input is $\vx$ and parameters $\Theta = (\theta_1,\dots, \theta_{K-1})$. The probability for bin $k< K$ is:
\begin{align*}
    \frac{\exp(\sum_{j=k}^{K-1} \theta_{j}^T\vx)}{1+\sum_{i=1}^{K-1}\exp(\sum_{j=i}^{K-1} \theta_{j}^T\vx)},
\end{align*}
and the probability for bin $K$ is :
\begin{align*}
\frac{1}{1+\sum_{i=1}^{K-1}\exp(\sum_{j=i}^{K-1} \theta_{j}^T\vx)}.
\end{align*}

\section{Experimental Details \label{appsec:experimentdetails}}

\subsection{Gamma Simulation \label{appsec:experimentgamma}}

We use a $4$-layer neural network of hidden-layer sizes $128, 64, 64$ units, with \texttt{ReLU} activations to parameterize the categorical and log-normal distributions. For categorical we use another linear transformation to map to 50 output dimensions. For the log-normal model, two copies of the above neural network are used,
one to output the location and the other to output the log of the log-normal scale parameter. For \gls{mtlr}, we use a linear transformation from covariates to 50 dimensions and use a softmax layer to output the probability for the 50 bins.  We use $0$ dropout, $0$ weight decay, learning rate 1e-3 and batch size 1000 for 100 epochs in this experiment. 

\subsection{Survial MNIST \label{appsec:experimentmnist}}

The model does not see the \acrshort{mnist} class and learns a distribution over times given pixels $\mbx_i$. We use a convolutional neural network.
We use several layers of 2D convolutions 
with a kernel of size $2$ and stride of size $1$. The sequence of channel numbers is $32, 64, 128, 256$ with the last layer containing scalars.   After each convolution, we use \texttt{ReLU}, then dropout, then size $2$ max pooling.

For categorical and log-normal models,
this CNN output is mapped through a three-hidden-layer \texttt{ReLU} neural network with hidden sizes $512, 1024, 1024$.
Between the fully connected layers, we use \texttt{ReLU} then dropout.
Again, with the log-normal, separate networks are used to output the location and log-scale. For \gls{mtlr}, the CNN output is linearly mapped to the 50 bins.
For categorical, we use $0.2$ dropout for uncensored and $0.1$ for censored. 
In \gls{mtlr}, we use dropout $0.2$. In lognormal, we use dropout $0.1$. 
We use weight decay 1e-4, learning rate 1e-3, and batch size 5000 for 200 epochs.

\subsection{MIMIC-III \label{appsec:experimentmimic}}

The input is high-dimensional (about $1400$) because it is a concatenated time series and because missingness masks are used. We use a 4-layer neural network of hidden-layer sizes $2048,1024,1024$ units with \texttt{ReLU}
activations.  For the categorical model, we use $B=20$ categorical output bins.  For the log-normal model, we use one three-hidden neural network of hidden-layer sizes $128, 64, 64$ units and an independent copy to output the location and log-scale parameters.  We use dropout $0.15$, learning rate 1e-3 and weight decay 1e-4 for 200 epochs at batch size 5000.

\subsection{The Cancer Genome Atlas, Glioma \label{appsec:experimentglioma}}

The Weibull model has parameters
scale and concentration.
The scale is set to
$\exp[\beta^\top \mbx]$ for regression parameters $\beta$, plus a constant
$1.0$ for numerical stability. We optimize the concentration parameter in $(1,2)$. The log-normal model is as described in the simulated gamma experiment,
except that it has two instead of three hidden layers, due to small data sample size. The categorical and \gls{mtlr} models
are also as described in the simulated gamma experiment, except that they have 20 instead of 50 bins, and are linear, again due to small data sample size.

We standardize this data and then clamp all covariates at absolute value $5.0$. For all models, we train for 10,000 epochs at learning rate 1e-2 with full data batch size 1201. We use 10 \gls{d-cal} bins for this experiment as studied in \cite{haider2020effective}, rather than the 20 bins used in all other experiments.

\section{Exploring Choice of $\gamma$ soft-indicator parameter
\label{appsec:gammahyperparameter}}
\begin{figure}[t]
  \centering 
  \includegraphics[width=\textwidth]{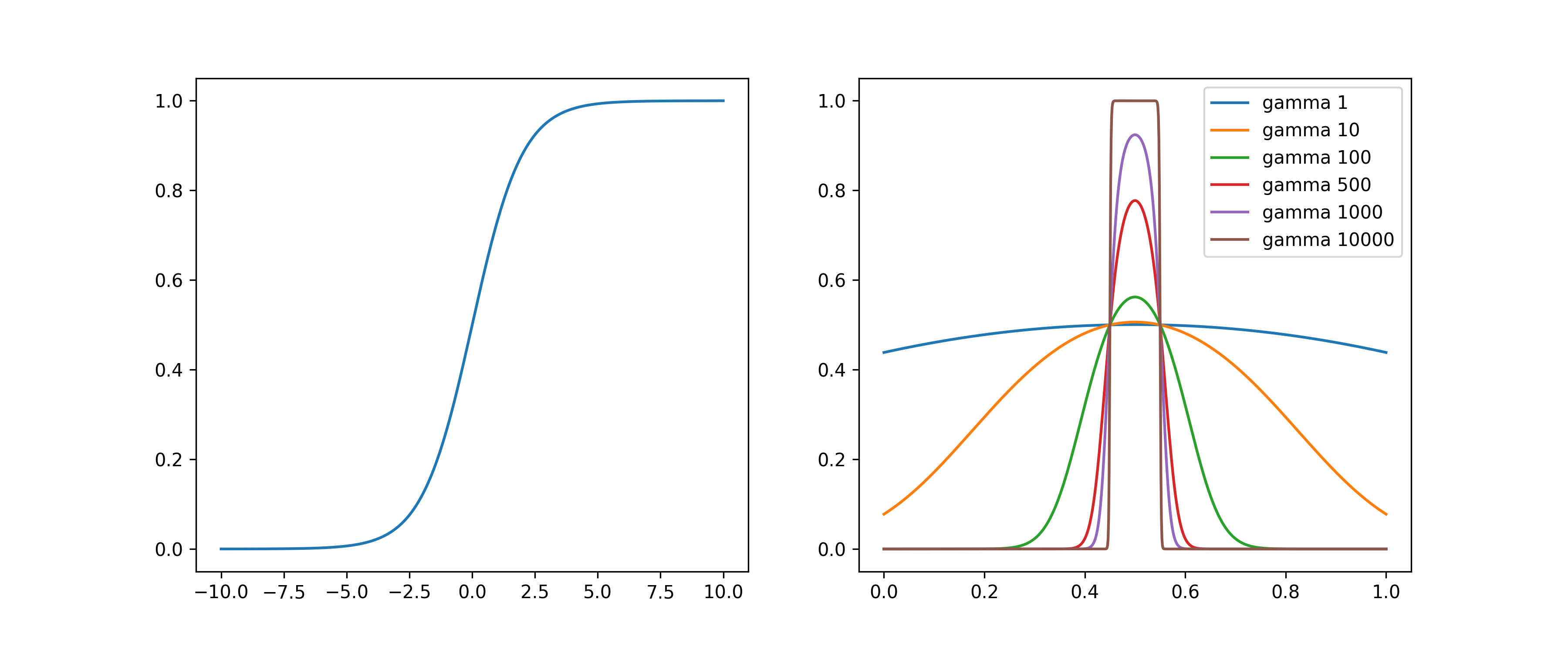} 
  \caption{Left: the sigmoid function. Right: choice of hyper-parameter gamma in soft indicator function for set $I=[0.45,0.55]$.}
  \label{fig:choiceofgamma} 
\end{figure}

 There is a trade-off in setting the soft membership parameter $\gamma$. 
Larger values approximate the indicator
function better, but can have worse gradients because the values
lie in the flat region of the sigmoid. See \Cref{fig:choiceofgamma} for an example of how gamma changes the soft indicator for a given set $I=[0.45,0.55]$.
We choose $\gamma=10000$ in all of the experiments and find that it
allows us to minimize exact \gls{d-cal-short}. We explore other choices in \Cref{tab:gammahyperparameter}.
We see the expected improvement in approximation as $\gamma$ increases. Then, as $\gamma$ gets too large, exact \gls{d-cal-short} stops improving as a function of $\lambda$.

\begin{table}[h]
    \centering
      \caption{Exact D-Cal, Soft-Dcal, and NLL at end of training, evaluated on training data for models trained with $\lambda=10$ and batch size $1,000$. Approximation improves as $\gamma$ increases. Gradients vanish when $\gamma$ gets too large. All experiments are better in calibration than the $\lambda=0$ MLE model,
    which has exact D-cal $0.09$. 
    }\label{tab:gammahyperparameter}
    \begin{tabular}{cccccccccc}\\\toprule  
        $\gamma$       & $10$  & $10^2$ & $10^3$ & $10^4$ & $10^5$  & $10^6$  & $10^7$ & $5 \times 10^7$
                             \\\midrule
        Exact D-Cal  & $0.2337$ & $0.0095$ &  $0.0079$ & $0.0039$ & $0.0025$  & $0.0014$  & $0.0015$ & $0.0048$
                              \\\midrule
        Soft D-Cal  & $0.4599$ & $0.0604$ &  $0.0074$ & $0.0039$ & $0.0025$  & $0.0014$  & $0.0015$ & $0.0048$
                            \\\midrule
        NLL  & $2.1180$ & $1.1362$ &  $1.0793$ & $1.2508$ & $1.6993$  & $2.3873$  & $2.6940$ & $3.4377$ \\
        \bottomrule
    \end{tabular}
\end{table}

\section{Exploring Slack due to Jensen's Inequality
\label{appsec:jensenslack}}

We trained the Categorical model on the gamma simulation data with $\gamma=10,000$ and batch size $10,000$ for all $\lambda$. The trained models are evaluated on the training set (size $100,000$) with two different test batch sizes, $500$ and $1000$. \Cref{tab:jensenslack} demonstrates that the upper-bounds for both batch sizes preserve model ordering with respect to exact \gls{d-cal}. The bound for batch size $10,000$ is quite close to the exact \gls{d-cal}.

\begin{table}[h]
\centering
        \caption{Slack in the upper-bound preserves modeling ordering with respect to exact \gls{d-cal}}\label{tab:jensenslack}
    \begin{tabular}{ccccc}\\ 
    \toprule
    $\lambda$  & Batch Size & Exact D-Cal  & Upper-bound \\\midrule
    $0$   & $500$ & $0.05883$  & $0.0605$ \\
           & $10000$ & ''      & $0.0589$  \\\midrule
           
     $1$   & $500$ & $0.02204$  & $0.0238$ \\
           & $10000$ & ''      & $0.0221$  \\\midrule
           
      $5$   & $500$ & $0.00963$  & $0.0114$ \\
           & $10000$ & ''      & $0.0097$  \\\midrule

      $10$   & $500$ & $ 0.00482$  & $0.0066$ \\
           & $10000$ & ''      & $0.0048$  \\\midrule
           
    $50$   & $500$ & $0.00040$  & $0.0021$ \\
           & $10000$ & ''      & $0.0004$  \\\midrule
           
        $100$   & $500$ & $0.00022$  & $0.0021$ \\
           & $10000$ & ''      & $0.0003$  \\\midrule
           
    $500$   & $500$ & $0.00015$  & $0.0020$ \\
           & $10000$ & ''      & $0.0002$  \\\midrule
           
   $1000$   & $500$ & $0.00006$  & $0.0019$ \\
           & $10000$ & ''      & $0.0001$  \\\midrule
    \end{tabular}

\end{table}

\section{\label{appsec:mod-soft-ind}Modification of soft indicator for the first and the last interval}
In our soft indicator,
\begin{align*}
    \zeta_\gamma(u; I) =  \text{Sigmoid}(\gamma(u-a)(b-u)) = \left(1 + \exp(-\gamma(u - a)(b - u))\right)^{-1}
\end{align*}
is a differentiable approximation for $\indicator{u \in [a,b]}$. When $b$ is the upper boundary of all the $u$ values, for example, 1 for \gls{cdf} values, the $b$ in the soft indicator can be replaced by any value that is greater than $b$. We use 2 to replace 1 for the upper boundary when $b=1$ in our experiments. Similarly we use $a=-1$ to replace $a=0$ for the lower boundary when $a=0$. 

Consider the term in our upper-bound (\cref{eq:upper-bound}) for the last interval $I=[a,b]$, where $b=1$, $\left(\frac{1}{M}\sum_i\zeta_\gamma(u_i; I)- |I|\right)^2$. The gradient of this term with respect to one \gls{cdf} value $u_i$ is:
\begin{align*}
    &\frac{d}{du_i} \left(\frac{1}{M}\sum_i\zeta_\gamma(u_i; I)- |I|\right)^2 
    \\ =&     \frac{d}{du_i}\left(\frac{1}{M}\sum_i
    \text{Sigmoid}(\gamma(u_i - a)(b - u_i))
    - |I|\right)^2\\
    &\Bigg[  \text { let }  A := 2/M * \left(\frac{1}{M}\sum_i
    \text{Sigmoid}(\gamma(u_i - a)(b - u_i)) - |I|\right) \Bigg]\\
    =& A \frac{d}{du_i}\text{Sigmoid}(\gamma (u_i - a)(b - u_i))
    \\
    =& A * -\frac{\exp(-\gamma(u_i - a)(b - u_i))}{(1+\exp(-\gamma(u_i - a)(b - u_i)))^2
    }\frac{d}{du_i}\left(-\gamma(u_i - a)(b - u_i)\right)\\
    =& A*\frac{\exp(-\gamma(u_i - a)(b - u_i))}{
    (1+\exp(-\gamma(u_i - a)(b - u_i)))
    ^2}*\gamma *(a+b-2u_i)
\end{align*}
If
\begin{align*}
    \frac{1}{M}\sum_i\zeta_\gamma(u_i; I)- |I| > 0,
\end{align*} 
then the fraction of points in the interval is larger than the size of the interval. We want to move the points out of the interval. In the last interval, in order to move points out of the interval, we can only make the values smaller, which means we want the gradient with respect to $u$ to be positive. (recall that we are moving in the direction of the negative gradient to minimize the objective). However, for points that are greater than $(a+b)/2$, the above gradient will be negative because term $(a+b-2u_i)$ is negative. This is not ideal. Changing the value $b$ from 1 to 2 can resolve the issue. Since \gls{cdf} values are all smaller than 1, $(a+b)/2$ will always be greater than $u$ if we use $b=2$ for the last interval. The above optimization issue only applies on the first and last interval because for intervals in the middle, we can move the points either to left or right to lower the fraction of points in the interval.

\newpage 

\section{Full Results: More Models and Choices of Lambda \label{appsec:fullresults}}

%CAT gamma uncensored interpolation
\begin{table}[h]
\centering
\caption{\label{tab:gammauncensoredfull} Gamma simulation, uncensored (full results)}
 \begin{tabular}{llllllllll}
\toprule
& $\lambda$&0&1&5&10&50&100&500&1000\\\midrule
Log-Norm & \acrshort{nll}
&0.381&0.423&0.507&0.580&0.763&0.809&0.870&0.882\\
\gls{nll} & \acrshort{d-cal-short}
&0.271&0.060&0.021&0.011&0.001&4e-4&1e-4&7e-5\\
& \acrshort{conc}
&0.982&0.955&0.931&0.908&0.841&0.835&0.809&0.802\\\midrule 
Log-Norm & \acrshort{nll}
&0.455&0.614&0.730&0.781&0.837&0.848&0.869&0.965\\
\gls{s-crps} & \acrshort{d-cal-short}
&0.055&0.014&0.004&0.002&2e-4&1e-4&1e-4&1e-4\\
& \acrshort{conc}
&0.979&0.975&0.968&0.959&0.940&0.931&0.864&0.811\\\midrule 
Cat-\acrshort{ni} & \acrshort{nll}
&0.998&1.042&1.129&1.197&1.788&2.098&3.148&3.688\\
& \acrshort{d-cal-short}
&0.074&0.023&0.008&0.005&4e-4&4e-4&2e-4&1e-4\\
& \acrshort{conc}
&0.986&0.986&0.985&0.985&0.973&0.960&0.877&0.748\\\midrule 
Cat-\acrshort{i} & \acrshort{nll}
&0.997&1.001&1.029&1.083&1.763&2.083&3.167&3.788\\
& \acrshort{d-cal-short}
&0.002&0.002&0.001&0.002&5e-4&5e-4&1e-4&1e-4\\
& \acrshort{conc}
&0.986&0.986&0.986&0.985&0.972&0.960&0.874&0.699\\\midrule 
\gls{mtlr}-NI & \acrshort{nll}
&1.287&1.409&1.589&1.612&2.356&2.590&3.267&3.509\\
 & \acrshort{d-cal-short}
&0.027&0.027&0.015&0.008&5e-4&2e-4&2e-4&2e-4\\
& \acrshort{conc}
&0.986&0.986&0.983&0.981&0.952&0.940&0.909&0.899\\\midrule 
\gls{mtlr}-I & \acrshort{nll}
&1.392&1.419&1.616&1.823&2.165&2.612&2.982&3.184\\
 & \acrshort{d-cal-short}
&0.048&0.034&0.017&0.009&7e-4&2e-4&1e-4&1e-4\\
& \acrshort{conc}
&0.986&0.986&0.982&0.980&0.958&0.934&0.918&0.917\\\bottomrule
\end{tabular}
\end{table}

\begin{table}[h]
\centering
\caption{\label{tab:gammacensoredfull} Gamma simulation, censored (full results). For categorical model with interpolation, the \gls{d-cal-short} is already very low at $\lambda=0$ so it is hard to optimize this one further.}

\begin{tabular}{llllllllll}
\toprule 
& $\lambda$ &0&1&5&10&50&100&500&1000\\ 
\midrule 
Log-Norm & \acrshort{nll} &-0.059&-0.049&-0.022&0.004&0.099&0.138&0.191&0.215\\
\gls{nll} & \acrshort{d-cal-short}&0.029&0.020&0.008&0.005&7e-4&2e-4&6e-5&7e-5\\
&\acrshort{conc}&0.981&0.969&0.950&0.942&0.927&0.916&0.914&0.897\\ \midrule 
Log-Norm & \acrshort{nll}&0.038&0.084&0.119&0.143&0.185&0.201&0.343&0.436\\
\gls{s-crps} & \acrshort{d-cal-short}&0.017&0.007&0.003&0.001&1e-4&1e-4&5e-5&8e-5\\
& \acrshort{conc}&0.982&0.978&0.971&0.963&0.952&0.950&0.850&0.855\\
\midrule
Cat-\acrshort{ni} & \acrshort{nll}&0.797&0.799&0.805&0.822&1.023&1.149&1.665&1.920\\
 & \acrshort{d-cal-short}
&0.009&0.006&0.003&0.002&3e-4&2e-4&6e-5&6e-5\\
& \acrshort{conc}
&0.987&0.987&0.987&0.987&0.982&0.976&0.922&0.861\\\midrule 
Cat-\acrshort{i} &\acrshort{nll}
&0.783&0.782&0.788&0.795&0.948&1.124&1.686&1.994\\
 & \acrshort{d-cal-short}
&7e-5 &1e-4&6e-5&8e-5&2e-4&2e-4&4e-5&6e-5\\
& \acrshort{conc}
&0.987&0.987&0.987&0.987&0.983&0.976&0.933&0.847\\\midrule 
\gls{mtlr}-NI & \acrshort{nll}
&0.873&0.875&0.875&0.977&1.271&1.412&1.747&1.900\\
 & \acrshort{d-cal-short}
&0.004&0.004&0.003&0.004&4e-4&2e-4&2e-4&2e-4\\
& \acrshort{conc}
&0.987&0.987&0.987&0.985&0.973&0.965&0.951&0.943\\\bottomrule
\gls{mtlr}-I & \acrshort{nll}
&0.829&0.830&0.866&0.981&1.266&1.414&1.762&1.912\\
 & \acrshort{d-cal-short}
&0.004&0.004&0.004&0.004&5e-4&1e-4&6e-5&7e-5\\
& \acrshort{conc}
&0.988&0.988&0.987&0.985&0.971&0.963&0.947&0.939\\
\bottomrule
\end{tabular}
\end{table}

\begin{table}[b]
\centering
\caption{\label{tab:mnistuncensoredfull} Survival-\acrshort{mnist}, uncensored (full results)}
 \begin{tabular}{llllllllll}
\toprule
& $\lambda$&0&1&5&10&50&100&500&1000\\\midrule
Log-Norm & \acrshort{nll}
&4.344&4.407&4.530&4.508&4.549&4.571&5.265&5.417\\
\gls{nll} & \acrshort{d-cal-short}
&0.328&0.104&0.018&0.020&0.011&0.010&0.005&0.005\\
& \acrshort{conc}
&0.886&0.867&0.754&0.759&0.725&0.713&0.541&0.509\\\midrule 
Log-Norm & \acrshort{nll}
&4.983&4.940&4.853&4.759&4.714&4.673&4.852&5.118\\
\gls{s-crps} & \acrshort{d-cal-short}
&0.212&0.132&0.081&0.059&0.020&0.007&0.003&0.003\\
& \acrshort{conc}
&0.889&0.878&0.866&0.861&0.873&0.873&0.820&0.798\\\midrule 
Cat-\acrshort{ni} & \acrshort{nll}
&1.726&1.730&1.737&1.755&1.824&1.860&2.076&3.073\\
 & \acrshort{d-cal-short}
&0.019&0.013&0.008&0.005&9e-4&9e-4&6e-4&3e-4\\
& \acrshort{conc}
&0.945&0.945&0.945&0.937&0.921&0.916&0.854&0.690\\\midrule 
Cat-\acrshort{i} & \acrshort{nll}
&1.726&1.731&1.735&1.741&1.782&1.809&1.953&2.157\\
 & \acrshort{d-cal-short}
&0.007&0.005&0.003&0.002&6e-4&3e-4&4e-4&3e-4\\
& \acrshort{conc}
&0.945&0.945&0.945&0.945&0.940&0.937&0.897&0.830\\\midrule 
\gls{mtlr}-\acrshort{ni} & \acrshort{nll}
&1.747&1.745&1.749&1.772&1.832&1.850&2.075&2.419\\
 & \acrshort{d-cal-short}
&0.018&0.014&0.008&0.004&0.001&0.001&8e-4&0.002\\
& \acrshort{conc}
&0.944&0.945&0.945&0.944&0.934&0.934&0.870&0.808\\\midrule 
\gls{mtlr}-I & \acrshort{nll}
&1.746&1.746&1.752&1.756&1.779&1.802&1.975&2.560\\
 & \acrshort{d-cal-short}
&0.005&0.004&0.003&0.002&5e-4&4e-4&8e-4&0.001\\
& \acrshort{conc}
&0.944&0.944&0.945&0.944&0.941&0.936&0.886&0.806\\\bottomrule
\end{tabular}
\end{table}

\begin{table}[h]
\centering
\caption{\label{tab:mnistcensoredfull} Survival-\acrshort{mnist}, censored (full results)}
 \begin{tabular}{llllllllll}
\toprule
& $\lambda$&0&1&5&10&50&100&500&1000\\
\midrule
Log-Norm & \acrshort{nll} &4.337&4.377&4.433&4.483&4.602&4.682&4.914&5.151\\
\gls{nll} & \acrshort{d-cal-short}
&0.392&0.074&0.033&0.020&0.008&0.005&0.005&0.007\\
& \acrshort{conc}
&0.902&0.873&0.829&0.794&0.728&0.696&0.628&0.573\\ \midrule 
Log-Norm & \acrshort{nll}
&4.950&4.929&4.873&4.859&4.672&4.749&4.786&4.877\\
\gls{s-crps} & \acrshort{d-cal-short}
 &0.215&0.122&0.071&0.051&0.018&0.010&0.002&9e-4\\
& \acrshort{conc}
&0.891&0.881&0.871&0.874&0.866&0.868&0.839&0.815\\ \midrule 
Cat-\acrshort{ni} & \acrshort{nll}
 &1.733&1.734&1.738&1.765&1.827&1.861&2.074&3.030\\
 & \acrshort{d-cal-short}
&0.018&0.014&0.008&0.004&8e-4&5e-4&5e-4&4e-4\\
& \acrshort{conc}
&0.945&0.945&0.944&0.927&0.920&0.919&0.862&0.713\\\midrule 
Cat-\acrshort{i} & \acrshort{nll}
&1.731&1.731&1.741&1.750&1.779&1.805&1.955&2.113\\
 & \acrshort{d-cal-short}
&0.007&0.006&0.003&0.002&3e-4&4e-4&4e-4&3e-4\\
& \acrshort{conc}
&0.945&0.944&0.945&0.945&0.942&0.938&0.901&0.843\\\midrule 
\gls{mtlr}-NI&  \acrshort{nll}
&1.126&1.118&1.125&1.136&1.174&1.193&1.350&1.482\\
& \acrshort{d-cal-short}
&0.021&0.017&0.012&0.009&0.006&0.006&0.006&0.007\\
& \acrshort{conc}
&0.958&0.960&0.961&0.960&0.949&0.943&0.897&0.880\\\midrule 
\gls{mtlr}-I & \acrshort{nll}
&1.126&1.118&1.125&1.136&1.174&1.193&1.350&1.482\\
& \acrshort{d-cal-short}
&0.021&0.017&0.012&0.009&0.006&0.006&0.006&0.007\\
& \acrshort{conc}
&0.958&0.960&0.961&0.960&0.949&0.943&0.897&0.880\\\bottomrule
\end{tabular}
\end{table}

\begin{table}[h]
\centering
\caption{\label{tab:newmimicfull} \acrshort{mimic-iii} length of stay (full results)}
 \begin{tabular}{llllllllll}
\toprule
& $\lambda$&0&1& 5 & 10& 50& 100&500&1000\\\midrule
Log-Norm   & \acrshort{d-cal-short}
&0.860&0.639&0.210&0.155&0.066&0.046&0.009&0.005\\
\gls{s-crps} & \acrshort{conc}
&0.625&0.639&0.577&0.575&0.558&0.555&0.528&0.506\\\midrule 
Cat-\acrshort{ni} & \acrshort{nll}
&3.142&3.177&3.101&3.167&3.086&3.088&3.448&3.665\\
 & \acrshort{d-cal-short}
&0.002&0.002&0.002&0.001&3e-4&2e-4&1e-4&1e-4\\
& \acrshort{conc}
&0.702&0.700&0.701&0.699&0.695&0.690&0.642&0.627\\\midrule 
Cat-\acrshort{i} & \acrshort{nll}
&3.142&3.075&3.157&3.073&3.002&3.073&3.364&3.708\\
 & \acrshort{d-cal-short}
&4-e4&3e-4&3e-4&3e-4&4e-4&1e-4&5e-5&4e-5\\
& \acrshort{conc}
&0.702&0.702&0.701&0.702&0.698&0.695&0.638&0.627\\\bottomrule
\end{tabular}
\end{table}

\begin{table}[h]
\centering
\caption{\label{tab:gbmlggfull} The Cancer Genome Atlas, glioma (full results)}
 \begin{tabular}{llllllllll}
\toprule
&$\lambda$&0&1&5&10&50&100&500&1000\\\midrule
Weibull & \gls{nll}
&4.436&4.390&4.313&4.292&4.441&4.498&4.475&4.528\\
& \acrshort{d-cal-short}
&0.035 &0.028&0.014&0.009&0.003&0.003&0.004&0.007\\
& \acrshort{conc}
&0.788&0.785&0.781&0.777&0.731&0.702&0.608&0.575\\\midrule
Log-Norm & 
 \gls{nll}
&14.187&6.585&4.841&4.639&4.181&4.181&4.403&4.510\\
\gls{nll} & \acrshort{d-cal-short}
&0.059 &0.024 &0.012&0.010&0.003&0.003&0.002&0.004\\
& \acrshort{conc}
&0.657&0.632&0.673&0.703&0.778&0.805&0.474&0.387\\ \midrule 
Log-Norm & \acrshort{nll}
&5.784&5.801&5.731&5.698&5.047&4.892&4.750&4.712\\
\gls{s-crps} & \acrshort{d-cal-short}
&0.258&0.2585&0.257&0.252&0.100&0.0702&0.044&0.025\\
& \acrshort{conc}
&0.798&0.798&0.798&0.810&0.568&0.507&0.420&0.363\\\midrule 
Cat-\acrshort{ni} & \acrshort{nll}
&1.718&1.742&1.746&1.758&1.800&1.799&1.810&1.826\\
 & \acrshort{d-cal-short}
&0.008&0.003&0.002&0.002&0.003&0.003&0.003&0.002\\
 & \acrshort{conc}
&0.781&0.771&0.775&0.775&0.765&0.765&0.758&0.748\\\midrule 
Cat-\acrshort{i} & \acrshort{nll}
&1.711&1.718&1.733&1.726&1.743&1.787&1.781&1.789\\
 & \acrshort{d-cal-short}
&0.003&0.001&8e-4&0.001&0.002&0.002&0.002&0.002\\
& \acrshort{conc}
&0.778&0.779&0.780&0.798&0.804&0.803&0.806&0.802\\\midrule 
\gls{mtlr}-NI & \gls{nll}
&1.624&1.620&1.636&1.636&1.666&1.658&1.748&1.758\\
 & \acrshort{d-cal-short}
&0.009&0.007&0.007&0.005&0.003&0.003&0.002&0.002\\
& \acrshort{conc}
&0.828&0.829&0.822&0.824&0.814&0.818&0.788&0.763\\\midrule 
\gls{mtlr}-I & \acrshort{nll}
&1.616&1.626&1.612&1.612&1.632&1.640&1.636&1.753\\
 & \acrshort{d-cal-short}
&0.003&0.003&0.002&0.001&0.001&0.001&9e-4&0.001\\
& \acrshort{conc}
&0.827&0.825&0.831&0.829&0.824&0.823&0.825&0.783\\\bottomrule
\end{tabular}
\end{table}

\end{document}